\documentclass[lettersize,journal]{IEEEtran}
\usepackage{amsmath,amsfonts}
\usepackage{array}
\usepackage[caption=false,labelfont=sf,textfont=sf]{subfig}
\usepackage{textcomp}
\usepackage{stfloats}
\usepackage{url}
\usepackage{verbatim}
\usepackage{graphicx}
\usepackage{cite}
\usepackage{amsmath}
\usepackage{algpseudocode}
\usepackage{booktabs}
\usepackage{multirow}

\usepackage{amssymb}
\usepackage{algorithm}

\usepackage{tikz}  

\graphicspath{{figures/}}

\newcommand{\changes}[1]{{\color{black}{#1}}}
\definecolor{COLORTABLE}{rgb}{0,0,0}

\begin{document}

\title{Core-Set Selection for \\Data-efficient Land Cover Segmentation}

\author{
Keiller Nogueira$^{1,*}$, Akram Zaytar$^{2,*}$, Wanli Ma$^{3,4,*}$, Ribana Roscher$^{5,6}$, Ronny Hänsch$^{7}$, Caleb Robinson$^{2}$, \\
Anthony Ortiz$^{2}$, Simone Nsutezo$^{2}$, Rahul Dodhia$^{2}$, Juan M. Lavista Ferres$^{2}$, Oktay Karakuş$^{3}$, Paul~L.~Rosin$^{3}$\\[1em]
$^{1}$ University of Liverpool, Liverpool, L69 7ZX, England, UK\\
$^{2}$ Microsoft AI for Good Research Lab\\
$^{3}$ Cardiff University, School of Computer Science and Informatics, Cardiff CF24 4AG, UK\\
$^{4}$ University of Cambridge, Cambridge, CB3 0FA, England, UK\\
$^{5}$ Forschungszentrum Jülich GmbH, Jülich, Germany\\
$^{6}$ University of Bonn, Bonn, Germany\\
$^{7}$ Department SAR Technology, German Aerospace Center (DLR), Germany \\

\thanks{\noindent Manuscript received April 19, 2021; revised August 16, 2021. * indicates equal contribution.}
}

\markboth{Journal of \LaTeX\ Class Files,~Vol.~14, No.~8, August~2021}%
{Shell \MakeLowercase{\textit{et al.}}: A Sample Article Using IEEEtran.cls for IEEE Journals}


\maketitle

\begin{abstract}
The increasing accessibility of remotely sensed data and their potential to support large-scale decision-making have driven the development of deep learning models for many Earth Observation tasks.
Traditionally, such models rely on large datasets. However, the common assumption that larger training datasets lead to better performance tends to overlook issues related to data redundancy, noise, and the computational cost of processing massive datasets.
Effective solutions must therefore consider not only the quantity but also the quality of data.
\changes{
Towards this, in this paper, we introduce six basic core-set selection approaches -- that rely on imagery only, labels only, or a combination of both -- and investigate whether they can identify high-quality subsets of data capable of maintaining -- or even surpassing -- the performance achieved when using full datasets for remote sensing semantic segmentation.
We benchmark such approaches against two traditional baselines on three widely used land-cover classification datasets (DFC2022, Vaihingen, and Potsdam) using two different architectures (SegFormer and U-Net), thus establishing a general baseline for future works.
Our experiments show that all proposed methods consistently outperform the baselines across multiple subset sizes, with some approaches even selecting core sets that surpass training on all available data.
Notably, on DFC2022, a selected subset comprising only 25\% of the training data yields slightly higher SegFormer performance than training with the entire dataset.
}
This result shows the importance and potential of data-centric learning for the remote sensing domain.
The code is available at~\url{https://github.com/keillernogueira/data-centric-rs-classification/}.
\end{abstract}


\section{Introduction}
The rapid advancements in satellite technologies have significantly enhanced accessibility to Earth observation data, opening new opportunities for a better understanding of the Earth’s surface~\cite{ban2015global}.
\changes{
This accessibility has naturally led to the development and training of several deep learning methods using increasingly larger labeled datasets~\cite{schmitt2023there}, often emphasizing label quantity over quality.
However, indiscriminately increasing dataset size does not necessarily translate into improved model performance.
}
This is because larger datasets require significant human effort or the integration of weak labels that, in turn, can lead to the introduction of noise, bias, and inaccurate annotations.
In general, the current assumption that more data inherently leads to better outcomes tends to overlook the complexities of data distribution~\cite{roscher2024better}, the potential for introducing biases and noise, spurious correlations, the energy consumption~\cite{lannelongue2021green}, and the computational resources required for processing, labeling, and storing vast datasets.
Therefore, effective solutions should consider not only the quantity but also the quality of data.

\changes{
One promising data-centric strategy that addresses these challenges is core-set selection, which aims to identify a small but informative subset of training examples that preserves the essential characteristics of the full dataset, while maintaining or even enhancing overall model performance. 
}
Such a paradigm can assist in several aspects, such as improved computational efficiency, cost-effective data handling, enhanced model performance, effective use of labeled data, or efficient labeling of unlabeled data~\cite{phillips2017coresets, ng2022unbiggen,jarrahi2023principles,aroyo2022data}.

\changes{
Existing work has explored core-set selection across various domains, employing distinct approaches, such as clustering algorithms and gradient approximation~\cite{chai2023efficient, santos2021quality, pooladzandi2022adaptive}.
Some works perform core-set selection before training the final machine-learning model~\cite{chai2023efficient,santos2021quality}, making them agnostic to the downstream model but limited to a single selection stage.
In contrast, others integrate core-set selection into the training process by updating the selected subset at each epoch~\cite{sener2018active, pooladzandi2022adaptive, mirzasoleiman2020coresets}, yielding model-specific optimization at the cost of increased computational overhead.
}
Importantly, while some of these works perform core-set selection for image classification, to the best of our knowledge, there have been no initiatives exploiting this paradigm for remote sensing image segmentation, which presents unique and complex challenges.


\changes{
To address this gap, in this paper, we propose to establish a general and comprehensive baseline for core-set selection in remote sensing image segmentation.
Towards this, we introduce and benchmark six basic \textbf{model-agnostic} core-set selection approaches for remote sensing image segmentation based on several distinct premises, that rely on imagery only, labels only, and a combination of both, and that can be readily adapted to different scenarios, applications, and modern architectures, including vision-language and foundation models.
}
Specifically, given the training instances (i.e., images and their corresponding segmentation masks), the proposed approaches rank the examples from most to least valuable for training, based on their representativeness according to specific criteria.
\changes{
We then leverage such rankings to select the most representative examples (core-set) and train the models accordingly, reducing the training time and improving overall effectiveness by filtering out non-representative and/or noisy examples.
}
The main contributions of this paper are the following:
\begin{itemize}
    \item Introduction of six core-set selection approaches for remote sensing image segmentation, each based on different premises; and
    \item \changes{A full set of experiments comparing these approaches against two common baselines on three widely used datasets using two different architectures, thus establishing a benchmark for future research in core-set selection.}
\end{itemize}

Overall, this work, an outcome of the Data-Centric Land Cover Classification Challenge of the Workshop on Machine Vision for Earth Observation and Environment Monitoring (MVEO) 2023, fills a critical gap in the literature and demonstrates the potential of core-set selection in advancing remote sensing image segmentation as well as data curation and labeling.

\changes{
The remainder of this paper is organized as follows: Section~\ref{sec:related_work} reviews related work, Section~\ref{sec:methods} introduces the six core-set selection methods, Section~\ref{sec:experimental_setup} describes the experimental setup, Section~\ref{sec:results} presents results and discussion, and Section~\ref{sec:conclusion} concludes the paper.
}


\begin{figure*}[th]
    \centering
    \includegraphics[width=0.75\textwidth]{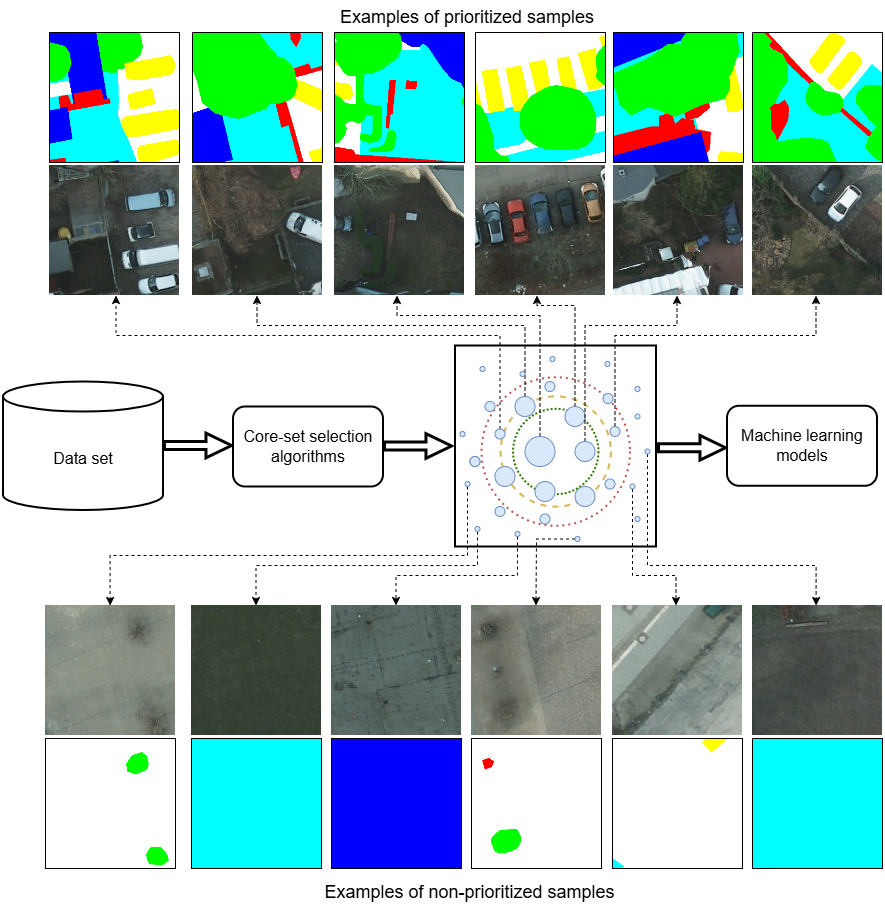}
    \caption{General overview of core-set selection. An input data set is first processed by a core-set selection algorithm that, based on some criteria, prioritizes certain examples over others (represented by the size of the blue circles). Based on this, it is possible to select the core-set data depending on the amount of data one would like to retain (illustrated by red, orange, and green dotted circles). Finally, the selected core-set is used to train a machine learning model, thus reducing the training time while maintaining, or even improving, task performance.}
    \label{fig:overview}
\end{figure*}

\section{Related work} \label{sec:related_work}


One of the major reasons to work with a core-set instead of the full data set is an improvement in computational efficiency.
Reducing the size of the dataset allows for quicker processing and experimentation.
This makes it possible to use complex machine-learning models without immense computational costs.
For this, oftentimes data-only techniques such as na\"ive random sampling are applied before model training, i.e., these methods do not need access to labels and are independent of the learning objective and application. 
An example is the identification of a subset that approximates the loss function of the whole dataset~\cite{chai2023efficient,pooladzandi2022adaptive,mirzasoleiman2020coresets}.
Furthermore, besides computational efficiency, smaller datasets reduce storage and maintenance costs, which is crucial when managing vast amounts of data from Earth observation systems.

Another reason is to improve model performance by enhancing the learning process and reducing overfitting through filtering out noisy or incorrect data points, thus creating cleaner datasets~\cite{li2021cleanml}.
In remote sensing, predominantly prior knowledge, label information, or an existing model is used to identify a clean core-set~\cite{ilyas2022machine,neutatz2021cleaning}.
Santos~\textit{et al.}~\cite{santos2021quality}, for example, use clustering for satellite time series to identify instances that are mislabeled or have low accuracy, with the goal of removing them from the training set to avoid a decrease in model performance.
Moreover, many methods have been developed for data with known sources of uncertainty, such as clouds~\cite{zhang2019coarse,li2020thin,ebel2022sen12ms}.
%
Furthermore, this reason is directly related to the enhancement of the accuracy and robustness of machine learning models by removing low-quality or redundant examples. Such models can perform as well as, or even better than, those trained on the full dataset (e.g.,~\cite{northcutt2019confidentlearning,citovsky2021batch,bahri2022margin}).

Another reason for core-set selection is to support clear and non-misleading explanations of a model and the data.
Generally, with the field of explainable machine learning, new methods are introduced to calculate importance scores, sensitivities, or contributions of features and interpret them as relevance~\cite{roscher2020explain}.
However, redundancies and correlations distort the derived insights, therefore they should be removed before interpreting and explaining the results. 
With the goal to analyze geospatial air quality estimations and the relevance of specific measurement locations, Stadtler \textit{et al.}~\cite{stadtler2022explainable} demonstrated that removing redundant examples only slightly decreases test accuracy, as these are not relevant for training. 

In general, the principles of core-set selection are closely related to areas like active learning~\cite{sener2018active,kim2022defense,zhang2021deep} and self-training~\cite{wei2023activeselfhar}.
For unlabeled datasets, core-set selection can guide efficient labeling by identifying the most representative examples.
This optimizes resource allocation for manual annotation - a common objective in active learning scenarios~\cite{sener2018active,kim2022defense,zhang2021deep}.
In case the data is already labeled, core-set selection can help prioritize the most informative examples.
This maximizes the use of labeled data and may reduce the need for further labeling efforts.

Overall, although most of the aforementioned works perform core-set selection for image classification, to the best of our knowledge, our work is the first research to design and benchmark core-set selection techniques specifically for remote sensing image segmentation.

\section{Core-set selection methods} \label{sec:methods}

Given a set of satellite images $\mathbf{X}$ and their corresponding label masks $\mathbf{M}$ (also known as segmentation maps), we introduce six basic core-set selection methods that assign an importance score, $s_i$, to the $i$-th instance $(X_i, M_i)$.
These scores range from 0 (least valuable for training) to 1 (most valuable for training).
\changes{
The aim of these methods is to rank the examples based on their informativeness, allowing us to select a subset of the data --- a \textit{core-set} --- that can achieve good model performance with reduced training time and data size.
}

\changes{
The proposed methods are categorized into three types: \textit{label-based} methods, which rely solely on the label masks $\mathbf{M}$; \textit{image-based} methods, which use information only from the input images $\mathbf{X}$; and \textit{combined} methods that integrate both sources.
A full summary of all proposed methods can be seen in Table~\ref{tab:methods_overview}.
}

For a given training budget $b$, the core-set consists of the top-$b$ examples ranked by their scores. Next, we describe the methods in detail.

\begin{table*}[htbp]
    \centering
    \caption{Overview of the six proposed core-set selection methods for remote sensing image segmentation.}
    \label{tab:methods_overview}
    \begin{tabular}{@{}>{\color{COLORTABLE}}l >{\color{COLORTABLE}}l >{\color{COLORTABLE}}l >{\color{COLORTABLE}}m{14em} >{\color{COLORTABLE}}l@{}}
        \toprule
        \multicolumn{1}{>{\color{COLORTABLE}}c}{\textbf{Method}} & \multicolumn{1}{>{\color{COLORTABLE}}c}{\textbf{Category}} & \multicolumn{1}{>{\color{COLORTABLE}}c}{\textbf{Input}} & \multicolumn{1}{>{\color{COLORTABLE}}c}{\textbf{Key Technique}} & \multicolumn{1}{>{\color{COLORTABLE}}c}{\textbf{Selection Criterion}} \\ 
        \midrule
        LC & Label-only & Masks ($\mathbf{M}$) & Entropy calculation & High entropy in class distribution indicates complex, informative masks \\
        FD & Image-only & Images ($\mathbf{X}$) & K-Means clustering & Diverse feature representations across clusters \\
        LC/FD & Hybrid & Both ($\mathbf{M,X}$) & Combined ranking & Diversity for small sets, complexity for larger sets (cutoff at $m=770$) \\
        FA & Image-only & Images ($\mathbf{X}$) & Feature activation statistics & High mean and standard deviation in ResNet-18 embeddings \\
        CB & Label-only & Masks ($\mathbf{M}$) & Iterative entropy maximization & Balanced class distribution across selected samples \\
        FA/CB & Hybrid & Both ($\mathbf{M,X}$) & Weighted ensemble ($\lambda=0.5$) & Combined feature activation and class balance scores \\
        \bottomrule
    \end{tabular}
\end{table*}



\subsection{Label Complexity (LC)}

\textit{LC} is a \textit{label-based} method that scores an data instance based on the \textbf{complexity} of its label mask $M_i$.
The underlying assumption is that examples with high-complexity label masks are more informative for training segmentation models.
It is important to emphasize that this approach does not explicitly guarantee representativeness but instead prioritizes complex label masks to potentially generate more informative training signals, which may lead to improved model performance.

The complexity of a label mask is quantified using the entropy of the class distribution in the label mask -- high entropy class distributions will include more and mixed classes, while low or zero entropy class distributions will be dominated by a single class. 
Precisely, for each instance $i$, we compute the score $s_i^\text{LC}$ based on the entropy $H(M_i)$ as:

\begin{equation} 
s_i^\text{LC} = H(M_i) = - \sum_{c=1}^{C} p_{i,c} \log_C (p_{i,c}), 
\end{equation}
where $C$ is the number of classes and $p_{i,c}$ is the proportion of pixels belonging to class $c$ in $M_i$. Classes that are labeled as ``unknown'' or ``ignored'' (in DFC2022) are excluded from this computation.

Higher scores correspond to examples that have a more uniform distribution of class labels with potentially more informative label masks, while low scores correspond to examples that are dominated by a single class.

\subsection{Feature Space Diversity (FD)}

\changes{
The \textit{FD} method is an \textit{image-based} approach that aims to select a diverse core-set of examples from $\mathbf{X}$ by leveraging feature embeddings extracted from a pre-trained deep learning model.
}

First, we embed each image, $X_i$, using a ResNet-18 model~\cite{he2016deep} pre-trained on ImageNet~\cite{deng2009imagenet}. We use the final feature representation layer (i.e. after spatial pooling) from the model, which results in a feature vector, $F_i \in \mathbb{R}^{512}$, that encodes higher-level semantic information per image.

Next, we group the feature embeddings into $K$ clusters using the K-Means algorithm. We search for a value of $K$ that minimizes the average Vendi score~\cite{dan2023vendi} across clusters. The Vendi score is a measure of diversity over a set of vectors -- by minimizing the average within-cluster diversity we ensure that an instance from that cluster is representative of the others.
Starting with $K=2$, we cluster all image embeddings with K-Means, measure the average per-cluster Vendi score, then increment $K$, and repeat until the change in the average Vendi score falls below a threshold of $\delta$ (we choose $\delta = 0.5\%$) for three iterations.

Given a clustering of the examples, we sequentially choose one instance randomly from within each cluster in a round-robin fashion (the first cluster is randomly selected from the set of $K$ clusters), resulting in ordered $K$-sized segments of cross-cluster examples. The first example has the highest importance score $s_i^\text{FD} = 1$, while the last selected example receives the lowest score $s_i^\text{FD} = 0$.

It is important to highlight here that this random selection step does not inherently produce a fixed ordering of scores.
However, this does not impact the primary objective of this method, which is to ensure diversity among the selected examples.

\subsection{Complexity/Diversity Hybrid (LC/FD)}

The \textit{LC/FD} method is a \textit{hybrid} method that combines the most important examples from the \textit{LC} and \textit{FD} methods, following an assumption that diversity is important when working with small datasets, while label complexity becomes increasingly important for medium to large datasets.

Specifically, the \textit{LC/FD} method is a \textit{hybrid} approach that uses the ranked lists of examples from \textit{LC} and \textit{FD}, denoted as $\mathcal{R}_{\text{LC}}$ and $\mathcal{R}_{\text{FD}}$, respectively.
A cutoff point $m$ is defined, and the hybrid ranking is constructed by taking the top $m$ examples from $\mathcal{R}_{\text{FD}}$, followed by all examples from $\mathcal{R}_{\text{LC}}$ that are not already included. This ensures that the selected core-set includes a mix of label-complex examples and feature-diverse instances.

\subsection{Feature Activation (FA)}

The \textit{FA} method is an \textit{image-based} approach that uses statistics from image embeddings created by a pre-trained neural network to rank examples.

First, a ResNet-18~\cite{he2016deep}, pre-trained on the ImageNet dataset, is used to extract the image embeddings (after the final spatial pooling layer), resulting in a feature vector, $F_i \in \mathbb{R}^{512}$, that encodes higher-level semantic information per image.

\changes{
By construction, all values in $F_i$ are non-negative due to the application of a ReLU activation within the network.
We assume that examples with high activation magnitudes (large mean) and significant variations across different dimensions (high standard deviation) in feature space are likely to carry more relevant information.
Accordingly, we compute the score $s_i^{\text{FA}}$ by combining the mean $\mu_i$ and the standard deviation $\sigma_i$ of the embedding vector $F_i$, with both quantities scaled to the interval $(0,1]$, as follows:
}


\begin{equation}
    s^{\text{FA}}_{i} = 1 - \left[ \frac{ \gamma_i - \min\limits_{F_j\in\mathbf{F}}[\gamma_j]}{\max\limits_{F_j\in\mathbf{F}}[\gamma_j]-\min\limits_{F_j\in\mathbf{F}}[\gamma_j]}\right],
\end{equation} 
where $\gamma_i = -(1-\mu_i) \cdot log(\sigma_i)$.

According to the aforementioned assumption, examples with low scores are likely to have lower diversity, contain more noise, etc., and are therefore likely to be less important for training.

\subsection{Class Balance (CB)}


Similar to the \text{LC}, the \textit{CB} method is a \textit{label-based} technique that aims to select a subset of examples with a uniform class distribution by using a time-efficient strategy that preprocesses and computes the class distribution of each image for subset selection.

\changes{
The method consists of $N$ steps, where $N$ refers to the number of examples in the dataset. In each step, the most suitable instance is selected from the dataset to ensure that the overall class distribution of the selected examples approaches a uniform distribution.
Specifically, the most suitable instance is the one that maximizes the entropy of the class distribution obtained by taking the union of the current core-set and the candidate example.
}


%
The order in which the examples are selected determines their importance score: the first instance is ranked as the most important and the last example is ranked as least important. Formally, let $r_i$ be the rank of instance $i$, then:

\begin{equation}
    s^{CB}_i = 1 - \frac{r_i}{N}.
\end{equation}

\subsection{Feature Activation/Class Balance Hybrid (FA/CB)}

The \textit{FA/CB} method is a \textit{hybrid} method that uses a weighted ensemble of importance scores calculated by the previous two methods, that is:

\begin{equation}
    s^{\text{FA/CB}}_i = \lambda \cdot s^{\text{FA}}_i + (1-\lambda) \cdot s^{\text{CB}}_i,
\end{equation}
where $\lambda$ is a trade-off weight between the two scores.

\section{Experimental Setup} \label{sec:experimental_setup}

\subsection{Datasets} \label{sec:datasets}

We test our approaches using three high-resolution datasets for semantic segmentation with remotely-sensed imagery as described below.

\subsubsection{IEEE GRSS Data Fusion Contest 2022 (DFC2022) Dataset} 

The DFC2022 dataset~\cite{hansch20222022} was released as part of the annual IEEE GRSS Data Fusion Contest.
This dataset consists of images gathered in and around 19 urban areas from different regions in France.
Each instance of this dataset contains a high-resolution RGB aerial image and its corresponding segmentation mask, both having approximately $2000\times2000$ pixels and a spatial resolution of 50 cm per pixel, along with a Digital Elevation Model (DEM) with $1000\times1000$ pixels at a spatial resolution of 100cm/pixel.
For our experiments, we resample the DEM data to match the dimensions of the image and mask.
The masks in this dataset contain 12 classes: ``urban fabric'', ``industrial, commercial, public, military, private and transport units'', ``mine, dump and construction sites'', ``artificial non-agricultural vegetated areas'', ``arable land'', ``permanent crops'', ``pastures'', ``forests'', ``herbaceous vegetation associations'', ``open spaces with little or no vegetation'', ``wetlands'', and ``water''.
Examples of this dataset are shown in Figure~\ref{fig:dfc_dataset}.

\newcommand{\exFigSize}{0.15}
\begin{figure}[!th]
	\centering
    \includegraphics[width=\exFigSize\textwidth]{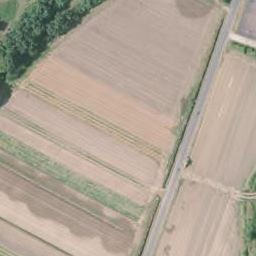}
    \includegraphics[width=\exFigSize\textwidth]{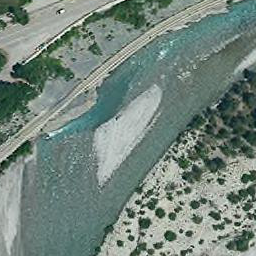}
    \includegraphics[width=\exFigSize\textwidth]{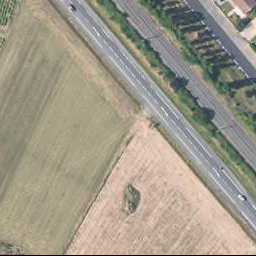}

    \vspace{2mm}
    \includegraphics[width=\exFigSize\textwidth]{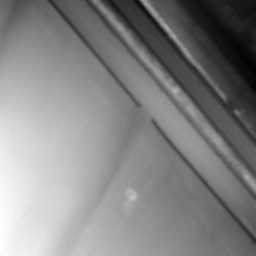}
    \includegraphics[width=\exFigSize\textwidth]{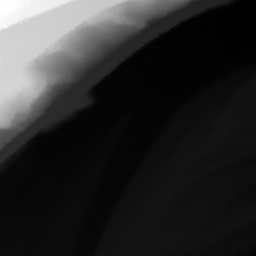}
    \includegraphics[width=\exFigSize\textwidth]{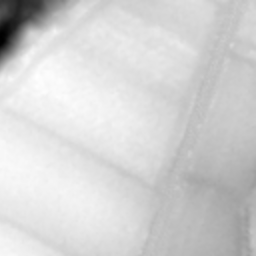}

    \vspace{2mm}
    \frame{\includegraphics[width=\exFigSize\textwidth]{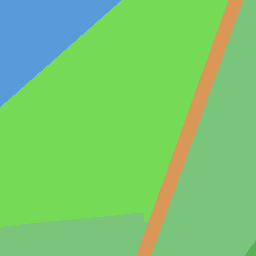}}
    \frame{\includegraphics[width=\exFigSize\textwidth]{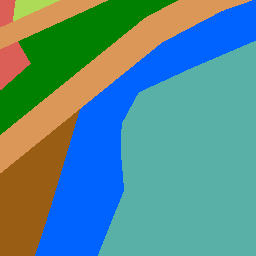}}
    \frame{\includegraphics[width=\exFigSize\textwidth]{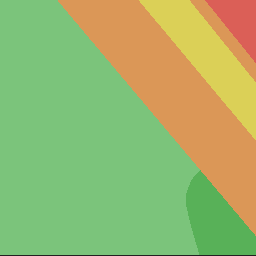}}

    \vspace{2mm}  
    \includegraphics[width=0.9\linewidth]{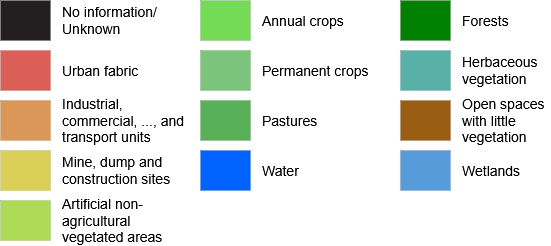}

	\caption{{\color{COLORTABLE} Example images (first row) of the DFC2022 dataset~\cite{hansch20222022}, their DEM data (second row), and the respective reference data (third row).}}
	\label{fig:dfc_dataset}
\end{figure}

\subsubsection{ISPRS Vaihingen and Potsdam Datasets} 

\begin{figure}[!th]
	\centering
    \includegraphics[width=\exFigSize\textwidth]{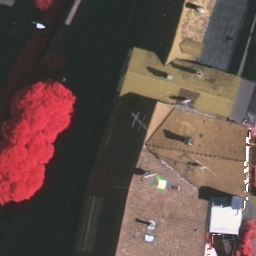}
    \includegraphics[width=\exFigSize\textwidth]{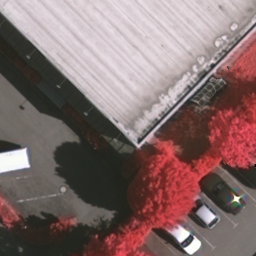}
    \includegraphics[width=\exFigSize\textwidth]{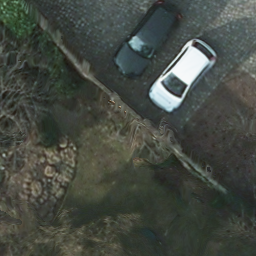}
    
    \vspace{2mm}
    \includegraphics[width=\exFigSize\textwidth]{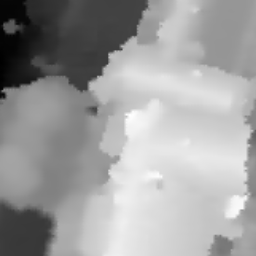}
    \includegraphics[width=\exFigSize\textwidth]{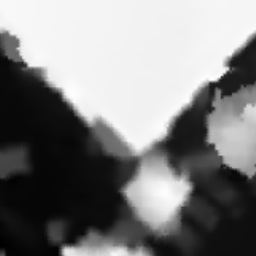}
    \includegraphics[width=\exFigSize\textwidth]{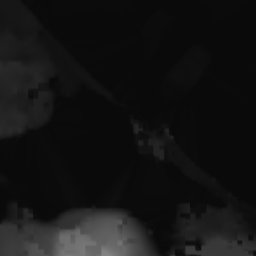}

    \vspace{2mm}
    \frame{\includegraphics[width=\exFigSize\textwidth]{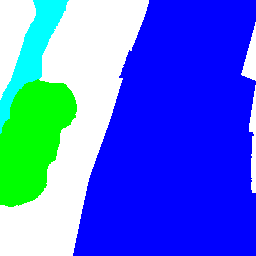}}
    \frame{\includegraphics[width=\exFigSize\textwidth]{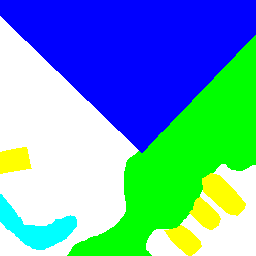}}
    \frame{\includegraphics[width=\exFigSize\textwidth]{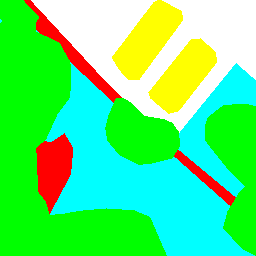}}

    \vspace{2mm}  
    \includegraphics[width=0.8\linewidth]{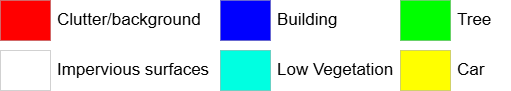}
	\caption{Example images (first row) of the Vaihingen and Potsdam datasets~\cite{isprs}, their DSM data (second row), and the respective reference data (third row).}
	\label{fig:vaihingen_dataset}
\end{figure}

The Vaihingen and Potsdam datasets~\cite{isprs} were released for the 2D semantic labeling contest of the International Society for Photogrammetry and Remote Sensing (ISPRS).
Both datasets consist of aerial imagery, Digital Surface Model (DSM) data, and label masks, as shown in Figure~\ref{fig:vaihingen_dataset}.
The Vaihingen dataset contains 33 patches, with an average size of $2494\times2064$ pixels. The aerial images have three bands (near-infrared, red, and green) with a spatial resolution of 9 cm per pixel.
The Postdam dataset contains 38 tiles of $6000 \times 6000$ pixels. The imagery consists of four bands (red, green, blue, and near-infrared) with a spatial resolution of 5 cm per pixel.
The label masks in both datasets contain six classes: ``impervious surfaces'', ``building'', ``low vegetation'', ``tree'', ``car'', and ``clutter/background''.

\subsection{Implementation Details} \label{sec:impl_details}

We preprocess the aforementioned datasets by tiling them into non-overlapping $256 \times 256$ patches that are used in all subsequent steps.

\changes{
We evaluate the effectiveness of the core-set selection methods using two popular segmentation networks: U-Net~\cite{ronneberger2015u} (with a ResNet-18~\cite{he2016deep} backbone), and SegFormer~\cite{xie2021segformer}.
The former is a convolutional segmentation model, while the latter is based on the transformer architecture.
For each experimental dataset, both models are trained exclusively on the selected core-set and evaluated on a held-out test set.
Importantly, our training and testing routine is fixed over the experiments, the only difference is in the subset (and size of subset) used to train the segmentation model.
}

\changes{
All proposed methods\footnote{The code is made publicly available at~\url{https://github.com/keillernogueira/data-centric-rs-classification/}} are implemented using Pytorch.
During training, we use the following hyper-parameters: 100 training epochs, AdamW~\cite{loshchilov2017decoupled} as optimizer, learning rate of 0.001, and batch size of 64.
}

For the \textit{LC/FD} method, we let $m = 770$ based on preliminary experiments with the DFC2022 dataset (we observed that \textit{FD} outperformed \textit{LC} for small subsets, while \textit{LC} added value for larger datasets).

For the \textit{FA/CB} method, we let $\lambda = 0.5$ to equally weight the importance from the \textit{FA} and \textit{FB} methods.

\subsection{Baselines} \label{sec:baseline}

\changes{
Given our general aim, we focus on model-agnostic core-set selection baseline methods, motivated by their high adaptability and usability, allowing them to be easily customized and applied across different downstream models and scenarios.
For all datasets, we compare the proposed techniques with two \textbf{traditional model-agnostic} baseline models:
}
(i) \textbf{Random}, which selects the core set uniformly at random, and 
(ii) \textbf{CoreSet}~\cite{sener2018active,citovsky2021batch,bahri2022margin}, which selects samples to optimally cover the embedding space.
Precisely, this approach iteratively expands the core-set by adding the data point that is farthest from its nearest neighbor in the current (core) set.
In this case, we used the Euclidean distance in the activations of the last spatial pooling layer of the ResNet-18~\cite{he2016deep}, similar to the approach used in the FA and FD methods.

\subsection{Experimental Protocol} \label{sec:protocol}

For the DFC2022 dataset, 90\% of the data originally released for the data fusion contest is made available to be ranked by the proposed core-set algorithms, while the remaining 10\% is used for validation.

For the Vaihingen and Potsdam datasets, we follow the standard protocol commonly exploited in the literature~\cite{nogueira2019dynamic}.
Specifically, for the Vaihingen dataset, 11 images originally released for the contest are made available for the proposed core-set techniques, and 5 images (with IDs 11, 15, 28, 30, 34) are employed for validation.
For the Potsdam dataset, 18 images released for the contest are made available for the proposed techniques, whereas 6 images (with IDs 02\_12, 03\_12, 04\_12, 05\_12, 06\_12, 07\_12) are used for validation.

For all datasets, the validation is only employed to assess the training of the U-Net models, after the selection of the core-set.
The final evaluation of the trained U-Net models uses the original test set of each dataset.
The overall performance of each method is measured by the mean Intersection over Union (mIoU) across all segmentation classes and averaged over three different model training runs.

\section{Experiments and Discussion} \label{sec:results}

\subsection{Quantitative Results} \label{sec:quant}

\changes{
To compare the performance of the introduced methods, we train and test both a U-Net and a SegFormer model (following the configuration described in Section~\ref{sec:impl_details}) on the top $1\%$, $5\%$, $10\%$, $25\%$, $50\%$, and $75\%$ ranked patches from each method.
The same experimental protocol is applied to the baseline approaches, with the additional inclusion of models trained on $100\%$ of the available training data, which serves as a robust reference baseline.
}
To account for potential variability due to randomness, three models are trained for each approach and subset size, using the same selected examples (per subset) and hyperparameters.
Finally, we used a paired t-test with $\alpha=0.05$ to evaluate statistically significant differences in results across methods.

\changes{
Tables~\ref{tab:results} and~\ref{tab:results_segformer} show the results for each method across both evaluated network architectures, the three datasets, and the different core-set sizes.
Overall, the proposed approaches consistently outperform the baselines across all datasets and architectures, including the setting where models are trained on 100\% of the available data.
Moreover, in most cases, the proposed approaches outperform both baselines even when using substantially fewer training examples.
For instance, on the Potsdam dataset, the U-Net model~\cite{ronneberger2015u} trained using only 25\% of the top-ranked samples selected by the FA/CB Hybrid approach achieves a higher mIoU score than both baselines trained on the top-ranked 50\% of the data. 
These results demonstrate that the proposed methods effectively identify representative and informative training samples, enabling more data-efficient and performant segmentation models.
}

\changes{
Furthermore, although the methods show varying degrees of effectiveness, the \textit{label-based} techniques outperformed the baselines at least once on all datasets across both networks.
Overall, these methods achieved the best overall performance (outperforming the baseline trained on 100\% of available data) in 3 dataset–model combinations, highlighting the importance of label diversity for effective core-set selection.
In comparison, the \textit{image-based} approaches yielded more moderate gains.
Although they outperformed the baselines in fewer cases, they still achieved the best performance in one dataset–model combination, specifically for the SegFormer model trained on the Vaihingen dataset, indicating that image-level representativeness alone can be beneficial in certain settings.
Finally, \textit{combined} approaches outperformed the baselines at least once across all datasets and both architectures, achieving the best overall results in 2 dataset–model combinations, showing that jointly exploiting label and image representativeness can be particularly beneficial in certain scenarios.
}

\changes{
We also observe that each dataset has a different IoU convergence rate.
DFC2022 has the most label noise (as evidenced by performance degradation of models in later splits), and the best-performing methods reached near-peak performance on both architectures by utilizing only 25-50\% of the data, noticeably outperforming training on 100\% of the data.
This rapid convergence suggests that, for datasets with specific characteristics (e.g., redundancy, noise), a relatively small core-set can be as effective, or even more effective, than the full dataset.
In contrast, on datasets with less label noise, such as Vaihingen and Potsdam, the performance continues to gradually improve on both architectures as the number of training examples increases.
However, even in these cases, the introduced methods are able to outperform the baseline trained on the full training set, demonstrating the importance of selecting a core set to deal with relevant issues such as noise, representativeness, and so on.
}
%

\begin{table*}[h]
\centering
\caption{Results (\% mIoU) of the \textbf{U-Net model}~\cite{ronneberger2015u} trained on subsets of varying sizes (1\%, 5\%, 10\%, 25\%, 50\%, 75\%, and 100\%) for the DFC2022, Vaihingen, and Potsdam datasets. Underlined values indicate the results that outperformed the corresponding baselines per training percentage (statistically significant paired t-test at $\alpha=0.05$). Bold values represent the best results overall for the dataset.}
\label{tab:results}
\footnotesize
\begin{tabular}{@{}l>{\color{black}}lrrrrrrc@{}}
\toprule
\multicolumn{1}{c}{\multirow{2}{*}{\textbf{Methods}}} & \multicolumn{7}{c}{\textbf{DFC 2022}} \\ \cmidrule(lr){2-9}
\multicolumn{1}{c}{} & \multicolumn{1}{c}{Category} & \textbf{1\%} & \textbf{5\%} & \textbf{10\%} & \textbf{25\%} & \textbf{50\%} & \textbf{75\%} & \textbf{100\%} \\ \midrule
Random & - & 10.42 $\pm$ 0.46 & 10.84 $\pm$ 0.15 & 11.55 $\pm$ 0.32 & 11.30 $\pm$ 0.66 & 12.41 $\pm$ 0.30 & 12.40 $\pm$ 0.36 & 12.71 $\pm$ 0.53  \\
CoreSet~\cite{sener2018active,citovsky2021batch,bahri2022margin} & - & 10.73 $\pm$ 0.23 & 11.47 $\pm$ 0.34 & 11.84 $\pm$ 0.42 & 11.47 $\pm$ 0.82 & 12.33 $\pm$ 0.27 & 12.23 $\pm$ 0.43 & - \\ \midrule
Label Complexity (LC) & Label-only & 10.42 $\pm$ 0.76 & \underline{12.44 $\pm$ 0.52} & \underline{12.83 $\pm$ 0.22} & \underline{12.68 $\pm$ 0.29} & \underline{\textbf{13.41 $\pm$ 0.61}} & 12.64 $\pm$ 0.35 & - \\
Feature Diversity (FD) & Image-only & 9.93 $\pm$ 0.24 & 11.87 $\pm$ 0.34 & 11.31 $\pm$ 0.29 & 11.61 $\pm$ 0.21 & 12.23 $\pm$ 0.08 & 12.00 $\pm$ 0.44 & - \\
LC/FD Hybrid & Both & 10.55 $\pm$ 0.16 & \underline{11.98 $\pm$ 0.27} & \underline{12.84 $\pm$ 0.33} & \underline{13.23 $\pm$ 0.12} & 12.62 $\pm$ 0.16 & 12.39 $\pm$ 0.30 & - \\
Feature Activation (FA) & Image-only & 8.67 $\pm$ 0.07 & 11.23 $\pm$ 0.65 & 11.14 $\pm$ 0.34 & 11.80 $\pm$ 0.42 & 12.66 $\pm$ 0.15 & 12.57 $\pm$ 0.11 & - \\
Class Balance (CB) & Label-only & 9.93 $\pm$ 1.36 & 10.38 $\pm$ 0.81 & 10.79 $\pm$ 0.20 & 11.52 $\pm$ 0.42 & 11.71 $\pm$ 0.21 & 12.27 $\pm$ 0.45 & - \\
FA/CB Hybrid & Both & 8.84 $\pm$ 0.53 & 10.44 $\pm$ 0.57 & 10.73 $\pm$ 0.50 & 11.90 $\pm$ 0.85 & 11.27 $\pm$ 0.32 & 12.22 $\pm$ 0.39 & - \\
\bottomrule
\end{tabular}%
\vspace{0.5cm}

\centering\small
\resizebox{\textwidth}{!}{%
\begin{tabular}{@{}l>{\color{black}}lccccccc@{}}
\toprule
\multicolumn{1}{c}{\multirow{2}{*}{\textbf{Methods}}} & \multicolumn{7}{c}{\textbf{Vaihingen}} \\ \cmidrule(lr){2-9}
\multicolumn{1}{c}{} & \multicolumn{1}{c}{Category} & \textbf{1\%} & \textbf{5\%} & \textbf{10\%} & \textbf{25\%} & \textbf{50\%} & \textbf{75\%} & \textbf{100\%} \\ \midrule
Random & - & 31.59 $\pm$ 1.59 & 39.89 $\pm$ 1.29 & 43.02 $\pm$ 1.48 & 53.72 $\pm$ 1.60 & 51.23 $\pm$ 1.89 & 57.86 $\pm$ 0.68 & 58.87 $\pm$ 0.57 \\
CoreSet~\cite{sener2018active,citovsky2021batch,bahri2022margin} & - & 29.71 $\pm$ 2.40 & 37.96 $\pm$ 2.30 & 41.39 $\pm$ 2.56 & 54.13 $\pm$ 1.68 & 55.00 $\pm$ 1.87 & 57.38 $\pm$ 0.96 & - \\
\midrule
Label Complexity (LC) & Label-only & \underline{34.98 $\pm$ 1.67} & \underline{42.58 $\pm$ 1.25} & 42.24 $\pm$ 1.39 & 55.14 $\pm$ 0.88 & 52.89 $\pm$ 3.85 & 58.35 $\pm$ 3.35 & - \\
Feature Diversity (FD) & Image-only & 31.40 $\pm$ 0.48 & 38.84 $\pm$ 1.81 & 42.99 $\pm$ 2.94 & 52.18 $\pm$ 3.12 & 53.19 $\pm$ 2.59 & 59.14 $\pm$ 1.67 & - \\
LC/FD Hybrid & Both & 23.52 $\pm$ 2.22 & 37.91 $\pm$ 1.30 & 41.22 $\pm$ 3.77 & 52.98 $\pm$ 2.72 & 50.44 $\pm$ 4.08 & 58.45 $\pm$ 0.42 & - \\
Feature Activation (FA) & Image-only & 27.53 $\pm$ 0.40 & 37.94 $\pm$ 4.58 & 43.59 $\pm$ 1.26 & 52.61 $\pm$ 0.11 & 56.34 $\pm$ 0.46 & 57.96 $\pm$ 0.19 & - \\
Class Balance (CB) & Label-only & 28.16 $\pm$ 5.08 & 32.22 $\pm$ 2.24 & \underline{45.65 $\pm$ 1.08} & 55.48 $\pm$ 1.61 & 54.64 $\pm$ 3.98 & \underline{60.54 $\pm$ 1.33} & - \\
FA/CB Hybrid & Both & 30.48 $\pm$ 5.86 & 40.69 $\pm$ 0.42 & 45.33 $\pm$ 1.90 & 56.77 $\pm$ 1.42 & 54.79 $\pm$ 2.97 & \underline{\textbf{61.27 $\pm$ 0.47}} & - \\
\bottomrule
\end{tabular}%
}
\vspace{0.5cm}

\centering\small
\resizebox{\textwidth}{!}{%
\begin{tabular}{@{}l>{\color{black}}lccccccc@{}}
\toprule
\multicolumn{1}{c}{\multirow{2}{*}{\textbf{Methods}}} & \multicolumn{7}{c}{\textbf{Potsdam}} \\ \cmidrule(lr){2-9}
\multicolumn{1}{c}{} & \multicolumn{1}{c}{Category} & \textbf{1\%} & \textbf{5\%} & \textbf{10\%} & \textbf{25\%} & \textbf{50\%} & \textbf{75\%} & \textbf{100\%} \\ \midrule
Random & - & 48.40 $\pm$ 1.97 & 68.42 $\pm$ 2.38 & 55.96 $\pm$ 2.81 & 72.04 $\pm$ 1.77 & 69.93 $\pm$ 1.28 & 78.75 $\pm$ 1.42 & 78.67 $\pm$ 1.40 \\
CoreSet~\cite{sener2018active,citovsky2021batch,bahri2022margin} & - & 52.84 $\pm$ 2.04 & 66.68 $\pm$ 3.55 & 67.13 $\pm$ 7.55 & 72.26 $\pm$ 2.04 & 68.74 $\pm$ 4.50 & 78.79 $\pm$ 1.40 & - \\ 
\midrule
Label Complexity (LC) & Label-only & \underline{56.45 $\pm$ 1.85} & 65.03 $\pm$ 1.28 & 66.48 $\pm$ 4.34 & 71.62 $\pm$ 4.31 & \underline{76.90 $\pm$ 3.48} & 80.39 $\pm$ 0.29 & - \\
Feature Diversity (FD) & Image-only & 49.06 $\pm$ 3.25 & 68.54 $\pm$ 0.95 & 66.08 $\pm$ 4.49 & \underline{74.67 $\pm$ 0.81} & 68.58 $\pm$ 2.03 & 79.54 $\pm$ 0.26 & - \\
LC/FD Hybrid & Both & 51.12 $\pm$ 3.52 & 69.39 $\pm$ 0.33 & 68.14 $\pm$ 6.37 & 72.96 $\pm$ 2.40 & 73.60 $\pm$ 5.87 & 80.98 $\pm$ 3.75 & - \\
Feature Activation (FA) & Image-only & 48.17 $\pm$ 1.38 & 67.30 $\pm$ 0.59 & 55.12 $\pm$ 1.66 & 70.37 $\pm$ 7.33 & 68.98 $\pm$ 0.99 & 80.04 $\pm$ 0.11 & - \\
Class Balance (CB) & Label-only & 48.36 $\pm$ 4.20 & 57.03 $\pm$ 5.13 & 64.35 $\pm$ 0.62 & 72.83 $\pm$ 0.07 & 73.89 $\pm$ 4.62 & 74.86 $\pm$ 5.36 & - \\
FA/CB Hybrid & Both & 52.51 $\pm$ 3.22 & 63.68 $\pm$ 0.85 & 62.64 $\pm$ 1.02 & 73.76 $\pm$ 0.23 & \underline{76.96 $\pm$ 1.21} & \underline{\textbf{81.28 $\pm$ 0.06}} & - \\
\bottomrule
\end{tabular}%
}
\end{table*}

\begin{table*}[h]
\centering
\caption{Results (\% mIoU) of the \textbf{SegFormer model}~\cite{xie2021segformer} trained on subsets of varying sizes (1\%, 5\%, 10\%, 25\%, 50\%, 75\%, and 100\%) for the DFC2022, Vaihingen, and Potsdam datasets. Underlined values indicate the results that outperformed the corresponding baselines per training percentage (statistically significant paired t-test at $\alpha=0.05$). Bold values represent the best results overall for the dataset.}
\label{tab:results_segformer}
\footnotesize
\resizebox{\textwidth}{!}{%
\begin{tabular}{@{}>{\color{COLORTABLE}}l >{\color{COLORTABLE}}l >{\color{COLORTABLE}}r >{\color{COLORTABLE}}r >{\color{COLORTABLE}}r >{\color{COLORTABLE}}r >{\color{COLORTABLE}}r >{\color{COLORTABLE}}r >{\color{COLORTABLE}}c@{}}
\toprule
\multicolumn{1}{>{\color{COLORTABLE}}c}{\multirow{2}{*}{\textbf{Methods}}} & \multicolumn{7}{>{\color{COLORTABLE}}c}{\textbf{DFC 2022}} \\ \cmidrule(lr){2-9}
\multicolumn{1}{c}{} & \multicolumn{1}{>{\color{COLORTABLE}}c}{Category} & \textbf{1\%} & \textbf{5\%} & \textbf{10\%} & \textbf{25\%} & \textbf{50\%} & \textbf{75\%} & \textbf{100\%} \\ \midrule
Random & - & 11.55 $\pm$ 0.26 & 12.48 $\pm$ 0.17 & 11.93 $\pm$ 0.37 & 12.27 $\pm$ 0.06 & 12.69 $\pm$ 0.50 & 11.87 $\pm$ 0.70 & 12.00 $\pm$ 0.45 \\
CoreSet~\cite{sener2018active,citovsky2021batch,bahri2022margin} & - & 11.84 $\pm$ 0.46 & 12.29 $\pm$ 0.55 & 12.02 $\pm$ 0.22 & 12.29 $\pm$ 0.37 & 12.23 $\pm$ 0.75 & 12.35 $\pm$ 0.24 & - \\
\midrule
Label Complexity (LC) & Label-only & 10.82 $\pm$ 0.53 & 12.21 $\pm$ 0.34 & 12.60 $\pm$ 0.55 & \underline{\textbf{13.01 $\pm$ 0.32}} & 12.64 $\pm$ 0.18 & 12.13 $\pm$ 0.11 & - \\
Feature Diversity (FD) & Image-only & 9.95 $\pm$ 0.58 & 11.44 $\pm$ 0.42 & \underline{12.43 $\pm$ 0.01} & 11.68 $\pm$ 0.54 & 11.62 $\pm$ 0.15 & 12.42 $\pm$ 0.25 & - \\
LC/FD Hybrid & Both & 11.03 $\pm$ 0.86 & 11.91 $\pm$ 1.07 & \underline{12.86 $\pm$ 0.06} & 11.97 $\pm$ 0.27 & 12.23 $\pm$ 0.23 & 11.82 $\pm$ 0.44 & - \\
Feature Activation (FA) & Image-only & 10.11 $\pm$ 1.14 & 10.90 $\pm$ 0.49 & 11.85 $\pm$ 0.57 & 12.59 $\pm$ 0.20 & 12.48 $\pm$ 0.06 & \underline{12.86 $\pm$ 0.26} & - \\
Class Balance (CB) & Label-only & 9.82 $\pm$ 0.42 & 10.16 $\pm$ 0.73 & 11.29 $\pm$ 0.45 & 11.39 $\pm$ 0.58 & 11.78 $\pm$ 0.76 & 12.33 $\pm$ 0.76 & - \\
FA/CB Hybrid & Both & 10.85 $\pm$ 1.03 & 10.35 $\pm$ 1.35 & 11.39 $\pm$ 0.87 & 11.08 $\pm$ 1.03 & 11.72 $\pm$ 0.65 & 12.36 $\pm$ 0.24 & - \\
\bottomrule
\end{tabular}%
}
\vspace{0.5cm}

\centering\small
\resizebox{\textwidth}{!}{%
\begin{tabular}{@{}>{\color{COLORTABLE}}l >{\color{COLORTABLE}}l >{\color{COLORTABLE}}c >{\color{COLORTABLE}}c >{\color{COLORTABLE}}c >{\color{COLORTABLE}}c >{\color{COLORTABLE}}c >{\color{COLORTABLE}}c >{\color{COLORTABLE}}c@{}}
\toprule
\multicolumn{1}{>{\color{COLORTABLE}}c}{\multirow{2}{*}{\textbf{Methods}}} & \multicolumn{7}{>{\color{COLORTABLE}}c}{\textbf{Vaihingen}} \\ \cmidrule(lr){2-9}
\multicolumn{1}{c}{} & \multicolumn{1}{>{\color{COLORTABLE}}c}{Category} & \textbf{1\%} & \textbf{5\%} & \textbf{10\%} & \textbf{25\%} & \textbf{50\%} & \textbf{75\%} & \textbf{100\%} \\ \midrule
Random & - & 12.02 $\pm$ 0.01 & 34.44 $\pm$ 8.05 & 51.96 $\pm$ 1.02 & 56.19 $\pm$ 2.25 & 56.01 $\pm$ 0.62 & 58.67 $\pm$ 1.01 & 57.71 $\pm$ 1.28 \\
CoreSet~\cite{sener2018active,citovsky2021batch,bahri2022margin} & - & 28.88 $\pm$ 2.58 & 42.34 $\pm$ 2.38 & 50.82 $\pm$ 2.38 & 57.77 $\pm$ 1.13 & 56.48 $\pm$ 1.55 & 60.09 $\pm$ 0.06 & - \\
\midrule
Label Complexity (LC) & Label-only & \underline{38.81 $\pm$ 0.05} & \underline{48.27 $\pm$ 0.30} & 51.62 $\pm$ 0.13 & 54.92 $\pm$ 2.07 & \underline{58.54 $\pm$ 0.16} & 58.02 $\pm$ 0.38 & - \\
Feature Diversity (FD) & Image-only & 29.36 $\pm$ 0.32 & 40.74 $\pm$ 2.53 & 50.81 $\pm$ 1.82 & 56.41 $\pm$ 1.57 & 56.72 $\pm$ 2.16 & 56.99 $\pm$ 2.81 & - \\
LC/FD Hybrid & Both & 30.65 $\pm$ 2.25 & 40.32 $\pm$ 6.68 & 51.30 $\pm$ 1.81 & 55.26 $\pm$ 0.78 & 55.36 $\pm$ 3.86 & 58.85 $\pm$ 1.22 & - \\
Feature Activation (FA) & Image-only & 34.12 $\pm$ 5.74 & 46.18 $\pm$ 2.41 & 51.41 $\pm$ 0.89 & 54.02 $\pm$ 0.77 & 55.60 $\pm$ 1.82 & \underline{\textbf{60.97 $\pm$ 0.46}} & - \\
Class Balance (CB) & Label-only & \underline{37.87 $\pm$ 1.98} & 41.63 $\pm$ 2.87 & 46.40 $\pm$ 4.24 & 54.31 $\pm$ 0.92 & 58.60 $\pm$ 1.91 & 58.72 $\pm$ 2.74 & - \\
FA/CB Hybrid & Both & \underline{34.40 $\pm$ 1.49} & \underline{48.72 $\pm$ 3.26} & 50.12 $\pm$ 3.61 & 58.34 $\pm$ 0.25 & 57.12 $\pm$ 3.32 & 60.39 $\pm$ 0.85 & - \\
\bottomrule
\end{tabular}%
}
\vspace{0.5cm}

\centering\small
\resizebox{\textwidth}{!}{%
\begin{tabular}{@{}>{\color{COLORTABLE}}l >{\color{COLORTABLE}}l >{\color{COLORTABLE}}c >{\color{COLORTABLE}}c >{\color{COLORTABLE}}c >{\color{COLORTABLE}}c >{\color{COLORTABLE}}c >{\color{COLORTABLE}}c >{\color{COLORTABLE}}c@{}}
\toprule
\multicolumn{1}{>{\color{COLORTABLE}}c}{\multirow{2}{*}{\textbf{Methods}}} & \multicolumn{7}{>{\color{COLORTABLE}}c}{\textbf{Potsdam}} \\ \cmidrule(lr){2-9}
\multicolumn{1}{c}{} & \multicolumn{1}{>{\color{COLORTABLE}}c}{Category} & \textbf{1\%} & \textbf{5\%} & \textbf{10\%} & \textbf{25\%} & \textbf{50\%} & \textbf{75\%} & \textbf{100\%} \\ \midrule
Random & - & 60.57 $\pm$ 0.05 & 70.62 $\pm$ 0.33 & 71.29 $\pm$ 3.63 & 68.88 $\pm$ 1.29 & 78.75 $\pm$ 0.14 & 76.62 $\pm$ 5.11 & 80.54 $\pm$ 2.07 \\
CoreSet~\cite{sener2018active,citovsky2021batch,bahri2022margin} & - & 60.56 $\pm$ 3.73 & 70.18 $\pm$ 0.35 & 69.55 $\pm$ 4.45 & 72.91 $\pm$ 3.81 & 78.94 $\pm$ 0.28 & 80.79 $\pm$ 2.25 & - \\
\midrule
Feature Diversity (FD) & Image-only & 56.07 $\pm$ 6.49 & 69.98 $\pm$ 0.30 & 69.91 $\pm$ 2.93 & 67.18 $\pm$ 1.72 & 76.82 $\pm$ 3.43 & 78.65 $\pm$ 7.72 & - \\
Label Complexity (LC) & Label-only & 60.75 $\pm$ 0.86 & 58.92 $\pm$ 2.19 & 71.87 $\pm$ 0.16 & 62.82 $\pm$ 1.42 & 78.13 $\pm$ 3.36 & \underline{83.78 $\pm$ 0.05} & - \\
LC/FD Hybrid & Both & 54.03 $\pm$ 0.87 & \underline{71.58 $\pm$ 0.64} & 67.48 $\pm$ 3.73 & 65.87 $\pm$ 1.58 & 78.43 $\pm$ 3.04 & \underline{83.83 $\pm$ 0.08} & - \\
Feature Activation (FA) & Image-only & 56.91 $\pm$ 0.43 & 60.72 $\pm$ 6.54 & 67.76 $\pm$ 2.49 & 63.71 $\pm$ 2.55 & 78.37 $\pm$ 1.31 & \underline{82.91 $\pm$ 0.02} & - \\
Class Balance (CB) & Label-only & 53.29 $\pm$ 0.70 & 55.33 $\pm$ 3.51 & 63.33 $\pm$ 2.63 & 67.71 $\pm$ 1.73 & 77.20 $\pm$ 3.87 & \underline{\textbf{83.96 $\pm$ 0.03}} & - \\
FA/CB Hybrid & Both & 59.27 $\pm$ 1.44 & 58.89 $\pm$ 4.63 & 65.98 $\pm$ 4.41 & 67.73 $\pm$ 3.01 & 76.56 $\pm$ 3.59 & \underline{83.91 $\pm$ 0.08} & - \\
\bottomrule
\end{tabular}%
}
\end{table*}

\changes{
In addition to performance gains, Table~\ref{tab:training_times} reports the training time per epoch (in seconds) of the U-Net model~\cite{ronneberger2015u}.
All experiments were conducted on a machine equipped with an Intel Xeon E5-2695 v4 (Broadwell) CPU, 128GB of RAM, and an NVIDIA P100 (Pascal) GPU with 16GB of memory, running CUDA 11.2 on Red Hat Enterprise Linux 8.2.
Training time is reported exclusively for the U-Net architecture as a representative model.
This is motivated by the fact that the relative reduction in computational cost is primarily driven by the size of the selected core-set and exhibits similar scaling behaviour across the different architectures.
Consequently, the observed trends generalize to the other model considered in this work.
Since all models are trained for a fixed number of 100 epochs, the reported per-epoch times allow us to directly quantify the computational savings enabled by core-set selection.
For example, when training the U-Net model~\cite{ronneberger2015u} on the DFC2022 dataset, the best-performing model (using only 50\% of the data) completed training more than 24 hours faster than the baseline trained on 100\% of the data, while achieving a superior test-set performance.
It is essential to observe that the computational overhead of the core-set selection process itself is minimal, amounting to at most the cost of a single training epoch using the full dataset, and is therefore \textbf{negligible} compared to the total training time.
In general, beyond reducing time, faster training enables the use of more training epochs and/or more extensive hyperparameter tuning, further supporting improved model optimization.
}

\begin{table}[htbp]
    \centering
    \caption{\changes{Training time per epoch (in seconds) of the U-Net model~\cite{ronneberger2015u} for different core-set sizes. The per-epoch training time is independent of the specific core-set selection method and depends solely on the size of the selected subset, given that the computational overhead of the core-set selection process itself is \textbf{negligible} compared to the full training procedure.}}
    \label{tab:training_times}
    \resizebox{\columnwidth}{!}{%
    \begin{tabular}{>{\color{COLORTABLE}}c>{\color{COLORTABLE}}r>{\color{COLORTABLE}}r>{\color{COLORTABLE}}r>{\color{COLORTABLE}}r>{\color{COLORTABLE}}r>{\color{COLORTABLE}}r>{\color{COLORTABLE}}r}
        \toprule
         Dataset & 1\% & 5\% & 10\% & 25\% & 50\% & 75\% & 100\% \\ \midrule
         DFC2022 & 40.65 & 107.34 & 194.62 & 428.14 & 810.97 & 1191.38 & 1981.11 \\
        Vaihingen & 4.10 & 7.92 & 8.70 & 11.91 & 20.11 & 28.94 & 38.47 \\
        Potsdam & 6.55 & 10.20 & 14.57 & 28.04 & 48.51 & 71.26 & 93.79 \\
        \bottomrule
    \end{tabular}
    }
\end{table}

\subsection{Qualitative Results} \label{sec:qual}

To facilitate the analysis and comparison of the proposed methods' outputs, we include visualizations of the average generated rankings, as can be seen in Figure~\ref{fig:results_patches}.
To generate these visualizations, the rankings produced by the different introduced methods are first averaged by patch position and then sorted, making the highest-ranking patches across the methods appear at the top.
Additionally, the standard deviation is calculated to capture the variability of the assigned ranks and provide insight into the approaches' consistency.
In addition to these visualizations, to allow for a better analysis, we also report the Kendall Tau correlations between each pair of methods in Figure~\ref{fig:correlation} and provide examples of the most and least frequently selected instances across all introduced methods in Figure~\ref{fig:top_bot_examples}.

\newcommand{\spaceFigs}{0.33\textwidth}
\begin{figure*}[t]
    \begin{center}
	\subfloat[DFC2022]{
        \includegraphics[width=\spaceFigs]{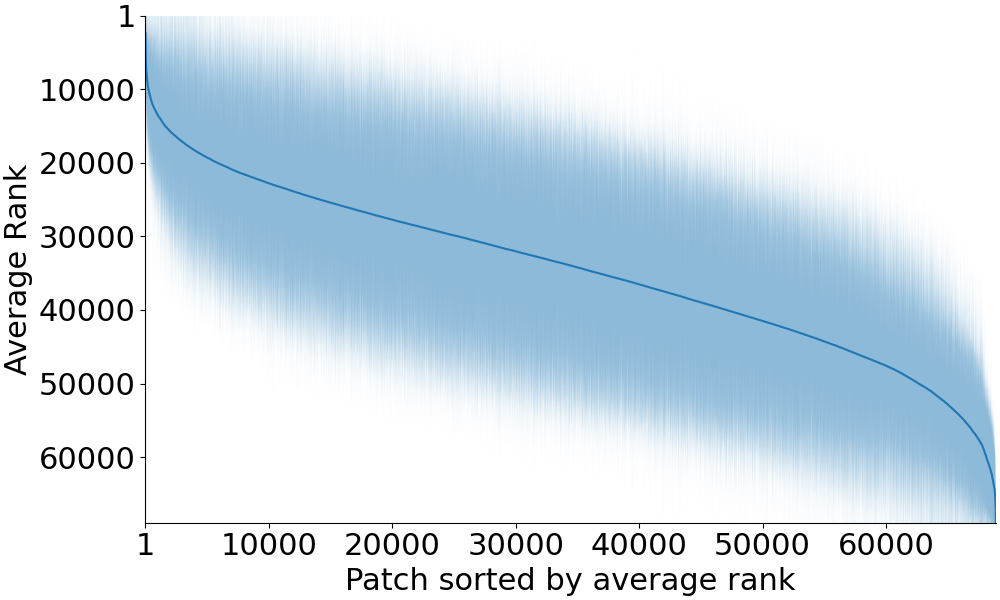}
	} 
	\subfloat[Vaihingen]{
		    \includegraphics[width=\spaceFigs]{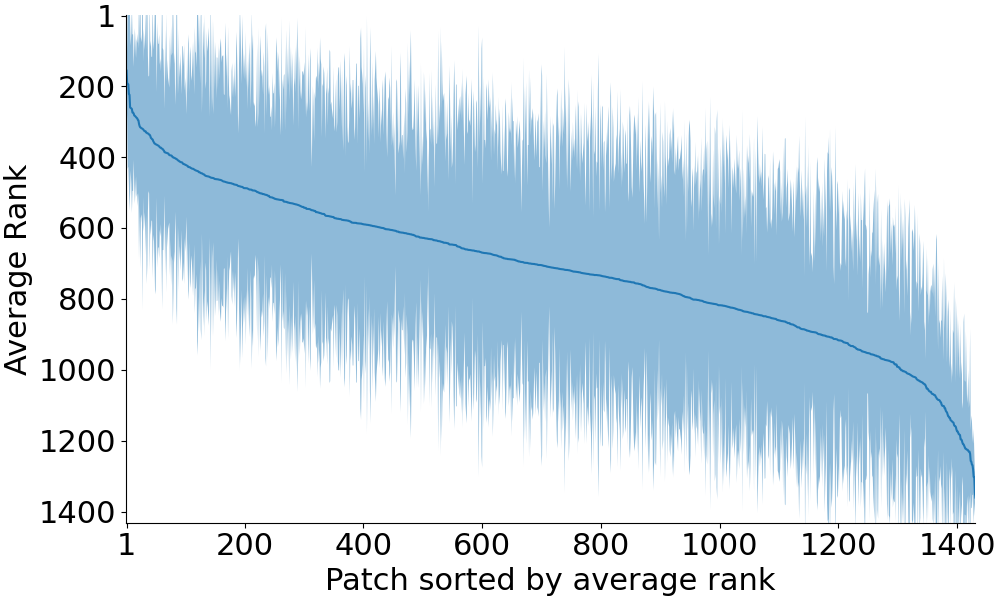}
	}
	\subfloat[Potsdam]{
        \includegraphics[width=\spaceFigs]{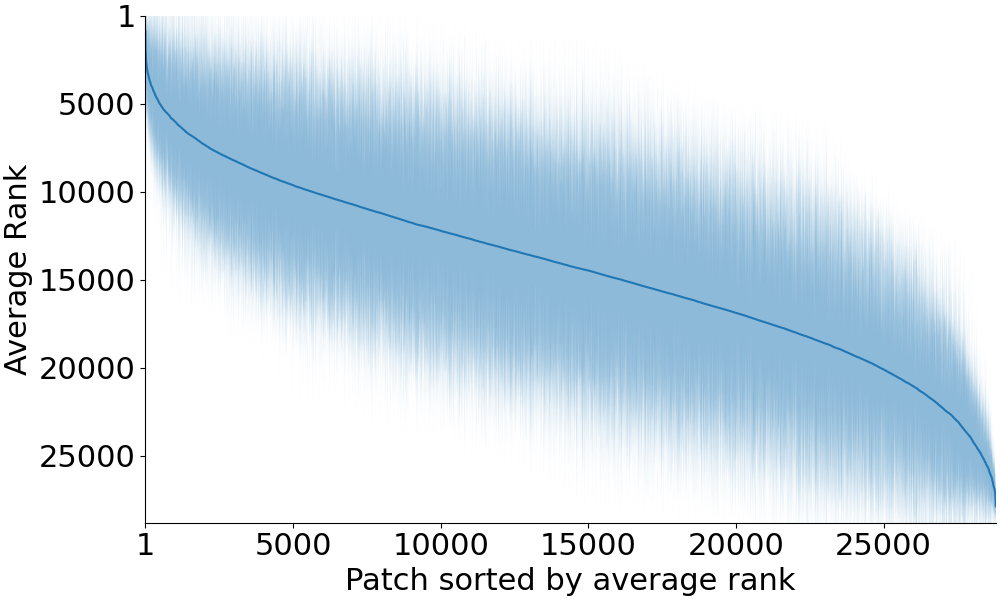}
	}
    \end{center}
	\caption{Visualizations of the proposed methods' rankings. The line represents the average rank position for each patch across all proposed approaches, while the shaded area represents the standard deviation.}
	\label{fig:results_patches}
\end{figure*}

\newcommand{\spaceFig}{0.31\textwidth}
\begin{figure*}[t]
    \begin{center}
	\subfloat[DFC2022]{
        \includegraphics[trim={4em 0 5em 0},clip,width=\spaceFig]{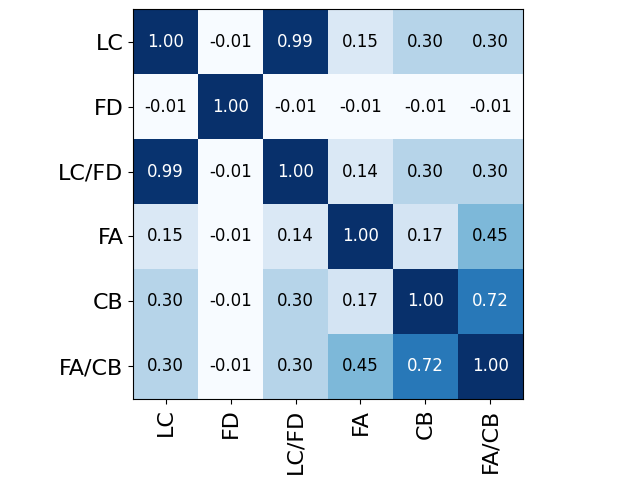}
	} 
	\subfloat[Vaihingen]{
		    \includegraphics[trim={4em 0 5em 0},clip,width=\spaceFig]{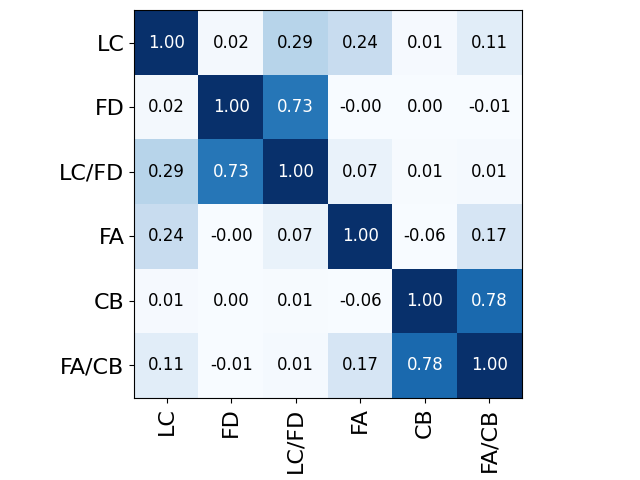}
	}
	\subfloat[Potsdam]{
        \includegraphics[trim={0 0 5em 0},clip,width=0.34\textwidth]{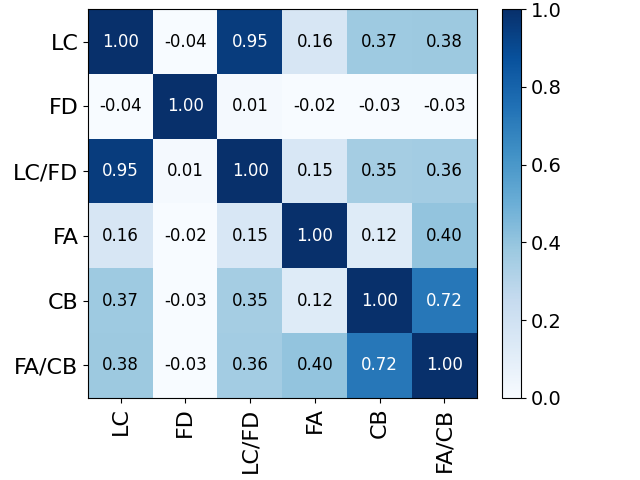}
	}
    \end{center}
	\caption{Correlation of methods according to Kendall Tau coefficient. A high correlation value means that the methods produce similar rankings.}
	\label{fig:correlation}
\end{figure*}

For all datasets, it is possible to observe that the approaches exhibit notable consistency, frequently assigning more importance to a specific (core) set of high-complexity examples that are consistently selected across the proposed methods.
Similarly, such approaches also tend to agree on the least important examples, assigning lower scores to low-complexity patches, indicating that the explored datasets contain a subset of non-representative or noisy instances that either contribute minimally to the overall performance or, in some cases, may even degrade it.
Furthermore, this level of agreement between the proposed techniques can be further observed in the correlation plots, wherein several methods show substantial correlation (particularly for the DFC2022 and Potsdam datasets), suggesting that they can capture underlying dataset patterns (such as the core sets).
Overall, the ability to select the core set, along with the identification of less valuable examples, highlights the robustness and efficiency of the proposed techniques in distinguishing between high- and low-quality data, thereby resulting in better performance and training time.
This is also qualitatively demonstrated in Figure~\ref{fig:top_bot_examples}: examples that are consistently highly ranked by the proposed methods show clear imagery with artifacts that are of a certain visual and semantic complexity (as illustrated by the corresponding label maps).
A few of these images allow learning the appearance and spatial relation of several classes at once.
On the other hand, images that are consistently rejected show very homogeneous scenes (such as large water bodies or parking lots) with neither much visual variation nor complex semantic content.

\newcommand{\exTopFigSize}{0.15\textwidth}
\begin{figure*}[t]
    \begin{center}
	\subfloat[DFC2022]{
        \begin{tabular}[b]{c}%
            \includegraphics[width=\exTopFigSize]{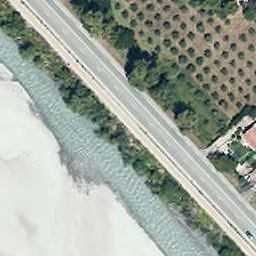} 
            \includegraphics[width=\exTopFigSize]{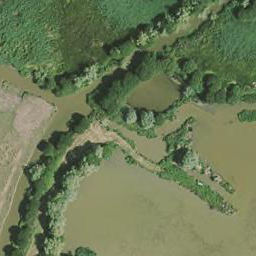}
            \includegraphics[width=\exTopFigSize]{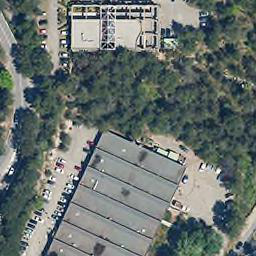} 
            \includegraphics[width=\exTopFigSize]{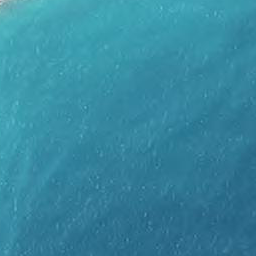}
            \includegraphics[width=\exTopFigSize]{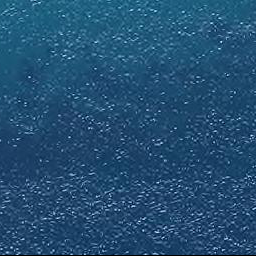}
            \includegraphics[width=\exTopFigSize]{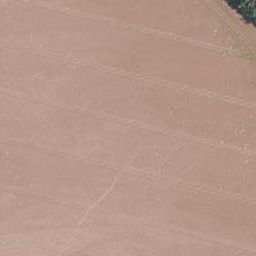} \\   
            \frame{\includegraphics[width=\exTopFigSize]{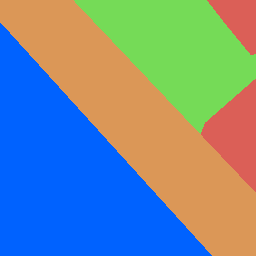}}
           \frame{\includegraphics[width=\exTopFigSize]{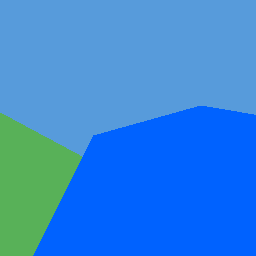}}
           \frame{\includegraphics[width=\exTopFigSize]{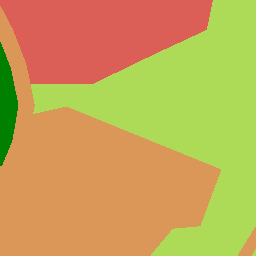}}
           \frame{\includegraphics[width=\exTopFigSize]{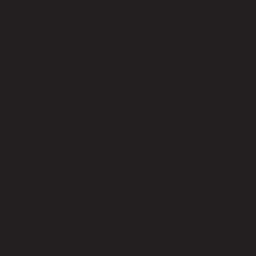}}
           \frame{\includegraphics[width=\exTopFigSize]{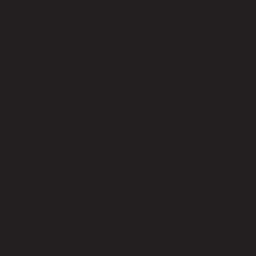}}
           \frame{\includegraphics[width=\exTopFigSize]{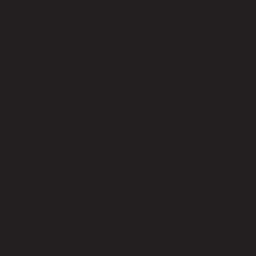}}
        \end{tabular}
	} \\
	\subfloat[Vaihingen]{
        \begin{tabular}[b]{c}%
            \includegraphics[width=\exTopFigSize]{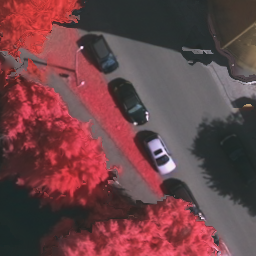} 
            \includegraphics[width=\exTopFigSize]{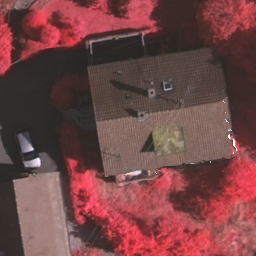}
            \includegraphics[width=\exTopFigSize]{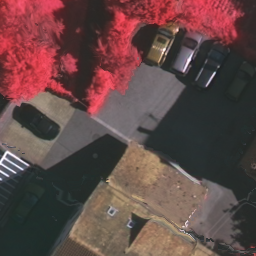} 
            \includegraphics[width=\exTopFigSize]{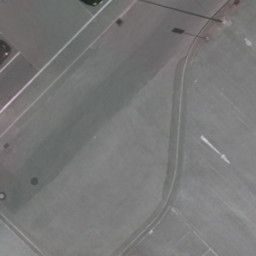}
            \includegraphics[width=\exTopFigSize]{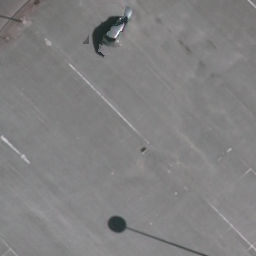}
            \includegraphics[width=\exTopFigSize]{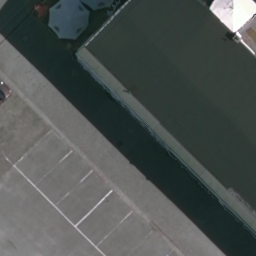} \\   
            \frame{\includegraphics[width=\exTopFigSize]{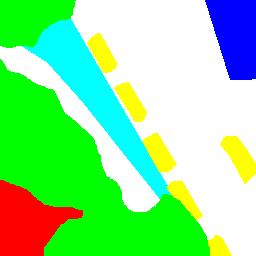}}
           \frame{\includegraphics[width=\exTopFigSize]{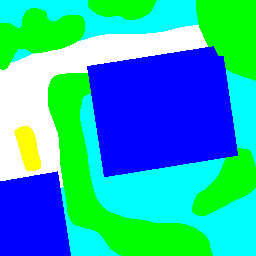}}
           \frame{\includegraphics[width=\exTopFigSize]{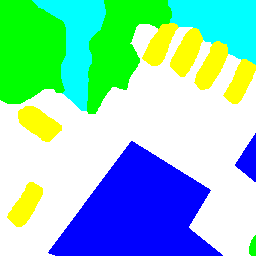}}
           \frame{\includegraphics[width=\exTopFigSize]{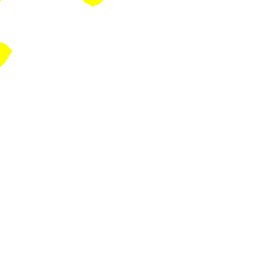}}
           \frame{\includegraphics[width=\exTopFigSize]{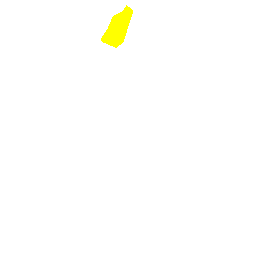}}
           \frame{\includegraphics[width=\exTopFigSize]{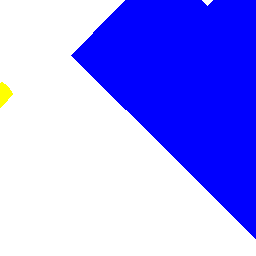}}
        \end{tabular}
	} \\
	\subfloat[Potsdam]{
        \begin{tabular}[b]{c}%
            \includegraphics[width=\exTopFigSize]{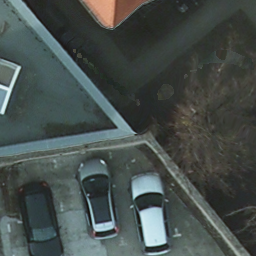} 
            \includegraphics[width=\exTopFigSize]{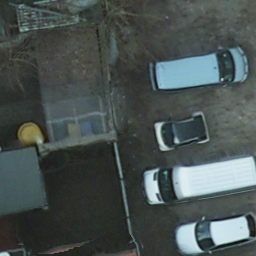}
            \includegraphics[width=\exTopFigSize]{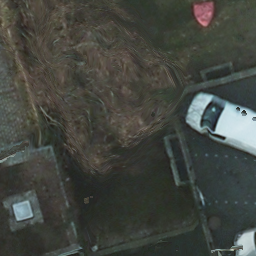} 
            \includegraphics[width=\exTopFigSize]{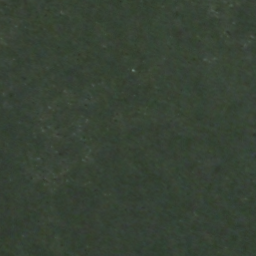}
            \includegraphics[width=\exTopFigSize]{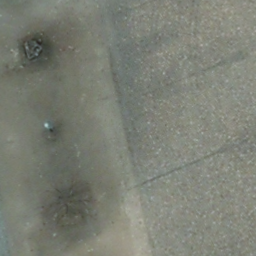}
            \includegraphics[width=\exTopFigSize]{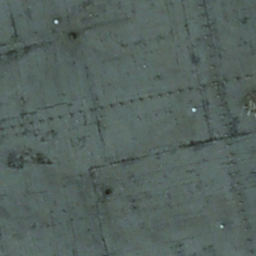} \\
            \frame{\includegraphics[width=\exTopFigSize]{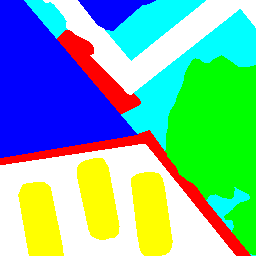}}
           \frame{\includegraphics[width=\exTopFigSize]{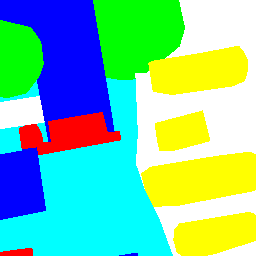}}
           \frame{\includegraphics[width=\exTopFigSize]{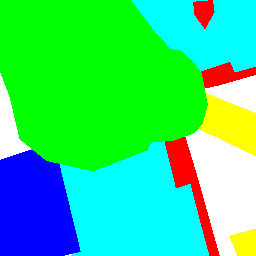}}
           \frame{\includegraphics[width=\exTopFigSize]{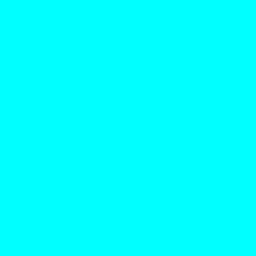}}
           \frame{\includegraphics[width=\exTopFigSize]{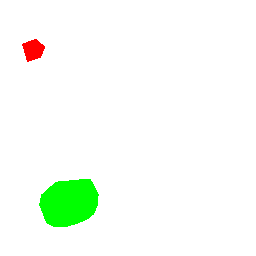}}
          \frame{\includegraphics[width=\exTopFigSize]{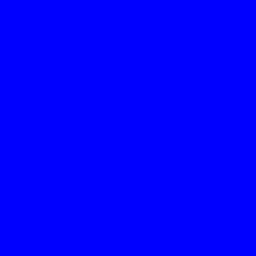}}
        \end{tabular}
	}
    \end{center}
	\caption{Examples of the highest-ranked (first three columns) and lowest-ranked instances (last three columns) considering the average ranking of all proposed methods.}
	\label{fig:top_bot_examples}
\end{figure*}

\section{Conclusions} \label{sec:conclusion}

In this paper, we introduce and systematically benchmark six basic core-set selection approaches for remote sensing image segmentation based on distinct premises - which rely on imagery only, labels only, or a combination of both - thereby establishing a general and comprehensive baseline for future works.
The proposed methods are able to consistently and effectively select the most important subset of examples (i.e., core-set), filtering out non-representative and noisy samples while preserving (or even improving) segmentation performance.

Extensive experiments are conducted using two different architectures (U-Net~\cite{ronneberger2015u} and SegFormer~\cite{xie2021segformer}) across three high-resolution remote sensing datasets with very distinct properties:
(i) IEEE GRSS Data Fusion Contest 2022 (DFC2022) dataset~\cite{hansch20222022}, consisting of very high-resolution visible spectrum images and Digital Elevation Model imagery, and
(ii) Vaihingen and Potsdam datasets~\cite{isprs}, both composed of high-resolution multispectral images and normalized Digital Surface Model data. 
This diverse experimental setup enables a robust assessment of the proposed methods across distinct settings.

Experimental results demonstrate the effectiveness and computational efficiency of the proposed core-set selection strategies, which consistently outperform traditional baselines.
Notably, on the DFC2022 dataset, the proposed approaches outperform the baseline trained on 100\% of the data while using only 25-50\% of the available examples.
Similarly, on the Vaihingen and Potsdam datasets, the same superior performance is achieved using just 75\% of the data.
These findings highlight that carefully selected core-sets can not only improve model performance but also substantially reduce training time and computational cost.


In summary, this work addresses a crucial gap in the literature and demonstrates the potential of core-set selection in advancing remote sensing image segmentation as well as data creation and labeling.
The presented conclusions open opportunities towards:
(i) the integration of core-set selection with other advanced techniques, such as self-supervised learning and foundation models, and
(ii) a more efficient and effective exploitation of both existing and new datasets for a better understanding of the Earth’s surface, an essential characteristic for most applications.





\bibliography{references}
\bibliographystyle{IEEEtran}



\end{document}